\documentclass[10pt,twocolumn]{ICCAS}
\usepackage{cite}
\usepackage{amsmath,amssymb,amsfonts}
\usepackage{algorithmic}
\usepackage{graphicx}
\usepackage{textcomp}
\usepackage{xcolor}
\usepackage{tabularx}
\usepackage{multirow}
\usepackage{url}
\usepackage{algorithm}
\usepackage{orcidlink}

\usepackage[caption=false,font=footnotesize]{subfig}

\usepackage{tikz}
\usetikzlibrary{arrows.meta,positioning,fit,calc}

\begin{document}

\title{Decoupled Object-Centric Video Understanding for Generating Robotic Manipulation Commands} 

\author{$^\dagger$Thanh Nguyen Canh${}^{1}$~\orcidlink{/0000-0001-6332-1002}, $^\dagger$Thanh-Tuan Tran${}^{2}$, Haolan Zhang${}^{1}$~\orcidlink{0009-0007-1742-3754}, Ziyan Gao${}^{1}$~\orcidlink{0000-0001-9948-7960}, \\ $^*$Xiem HoangVan${}^{2}$~\orcidlink{0000-0002-7524-6529}, and ${}^{*}$Nak Young Chong${}^{1,3}$~\orcidlink{0000-0001-5736-0769}}

\affils{${}^{1}$School of Information Science, Japan Advanced Institute of Science and Technology, Nomi, 923-1211, Ishikawa, Japan. \\ 
${}^{2}$University of Engineering and Technology, Vietnam National University, 10000, Hanoi, Vietnam. {\small$^\dagger$Equal contribution}\\ 
${}^{3}$Department of Robotics, Hanyang University, Ansan, 15588, Gyeonggi, Korea. 
{\small${}^{*}$ Corresponding author}}

\abstract{
Translating video demonstrations into executable robot commands remains challenging because existing methods often fail to identify which objects are functionally involved in the demonstrated action. As a result, they may generate commands that are linguistically plausible but operationally ambiguous. We propose an object-centric video understanding framework that decouples action recognition from object identification to generate precise, grammar-free manipulation commands. Our approach integrates Temporal Shift Modules (TSM) for efficient spatio-temporal action classification with a novel \textbf{Object Selection} algorithm that identifies task-relevant objects through trajectory-based role classification, blur detection, and overlap minimization. The selected objects are then processed by Vision-Language Models (VLMs) for robust category recognition and zero-shot generalization. Evaluated on a modified Something-Something V2 dataset, our method achieves 86.79\% action classification accuracy and BLEU-4 scores of 0.337 on standard objects and 0.261 on novel objects. These results improve over the strongest task-specific baseline by 80.2\% and 143.9\%, respectively. Larger gains are observed in METEOR and CIDEr, reaching 157.9\% and 171.7\% on novel objects. Across all semantic metrics, our approach consistently outperforms task-specific methods and remains competitive with, or surpasses, large general-purpose VLMs while retaining a modular, object-centric design.
}


\maketitle

\section{Introduction}
\label{sec:intro}

General video captioning methods~\cite{venugopalan2015sequence} focus mainly on global scene understanding and therefore lack the object-level precision required for robotic execution. In manipulation tasks, a robot must determine both \textit{which} action to perform and \textit{which} object to manipulate. Existing video-to-command approaches partially address this challenge. V2CNet~\cite{nguyen2018translating} improves action prediction through an auxiliary action classification branch, while Watch-and-Act~\cite{yang2023watch} introduces Visual Change Maps (VCM) to localize interaction regions through frame differencing. However, identifying task-relevant objects remains a fundamental difficulty. Specifically, the system must distinguish objects that are directly involved in the manipulation from irrelevant objects and incidental changes. Without this distinction, a generated command may contain the correct actions but refer to the wrong object, particularly in cluttered scenes or under occlusion.

Temporal modeling remains essential for action understanding. Conventional approaches, such as 3D CNNs~\cite{tran2015learning} and two-stream networks~\cite{simonyan2014two}, improve temporal reasoning but often require substantial computational cost. The Temporal Shift Module (TSM)~\cite{lin2019tsm} provides an efficient alternative by enabling temporal feature exchange without adding parameters. It has also demonstrated strong performance on fine-grained interaction datasets such as Something-Something-V2~\cite{goyal2017something}, whose temporal reasoning requirements are similar to those of manipulation action recognition.

For object identification, temporal modeling alone is insufficient. Recent Vision-Language Models (VLMs)~\cite{liu2023visual, li2023blip2} provide strong semantic understanding and zero-shot generalization, and they have been increasingly used in robotics for semantic grounding and task planning~\cite{shridhar2022cliport}. However, when VLMs are applied directly to full video frames, they must identify relevant objects while ignoring scene distractors. This makes object references inconsistent in cluttered manipulation scenarios. Therefore, action recognition and object identification require different processing strategies. Action recognition depends on temporal feature aggregation, whereas object identification requires localized, instance-level visual analysis.

To address these limitations, we propose an object-centric video understanding framework that decouples action recognition from object identification. The framework combines TSM-based action understanding with a trajectory-guided Object Selection module. This module identifies task-relevant objects through role assignment and quality filtering, and a VLM then recognizes the selected object categories for zero-shot generalization. The resulting action-object commands provide a structured and unambiguous interface for downstream robotic execution. The main contributions of this work can be summarized as follows:
\begin{figure*}[!ht]
\centering
\begin{tabular}{cc}
\includegraphics[width=1\textwidth]{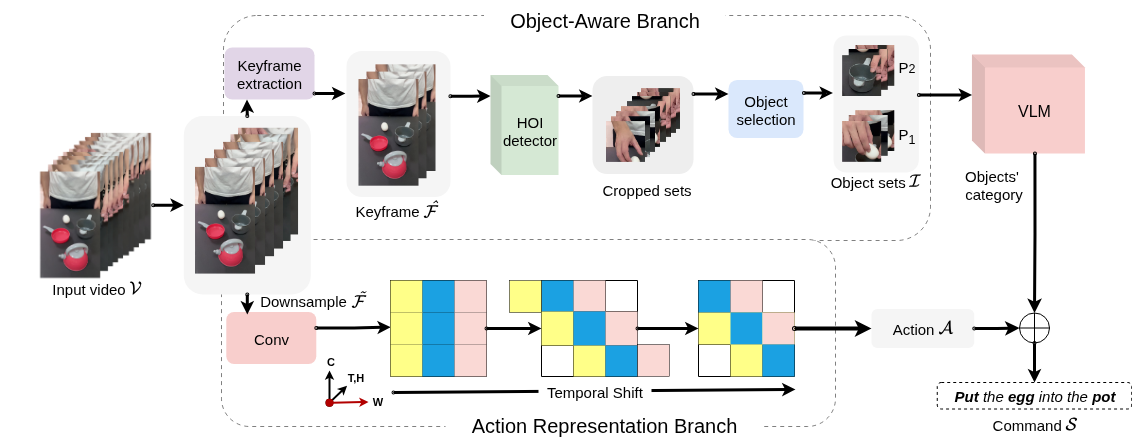} &
\end{tabular}
\caption{Proposed framework architecture. The Action Representation Branch applies a ResNet backbone with Temporal Shift Modules (TSM), followed by temporal pooling and a classifier. The Object-Aware Branch extracts keyframes, detects and tracks interacting objects, selects task-relevant instances through role and quality filtering, and uses a VLM for category recognition.}
\label{fig:architecture}
\end{figure*}

\begin{itemize}
\item A decoupled recognition-identification framework that separates action classification from object identification, enabling independent design and evaluation of each component.

\item  An object-centric selection pipeline that classifies objects into functional roles based on motion trajectories, then selects optimal instances through blur detection and overlap minimization.

\item An integration of a Vision-Language model for object identification, enabling generalization to novel categories without retraining.
\end{itemize}

\section{Proposed Method}~\label{sec:method}

\subsection{Framework Overview}

Given an input video $\mathcal{V}$ of a human manipulation demonstration, our goal is to generate a structured command $S$ that specifies the action and the involved objects. We decompose this task into two parallel branches: an Action Representation Branch and an Object-Aware Branch (Fig.~\ref{fig:architecture}). The Action Representation Branch predicts the manipulation action $\mathcal{A}$ from the downsampled video by modeling spatio-temporal dynamics (Sec. 3.2). The Object-Aware Branch localizes and identifies the task-relevant objects $\mathcal{I}$ using keyframe selection, hand--object interaction cues, tracking, and VLM-based recognition (Sec. 3.3). The raw video frames $\mathbf{F} = \{f_1, f_2, \ldots, f_n\}$ are first downsampled to $\tilde{\mathbf{F}} = \{\tilde{f}_1, \tilde{f}_2, \ldots, \tilde{f}_m\}$, where $m < n$. This step removes redundant frames while retaining sufficient temporal information. Because the action branch operates at a fixed sampling rate, downsampling also reduces test-time frame-rate variation that could shift the input distribution. Following standard formulations~\cite{venugopalan2015sequence}, command generation is posed as:

\begin{equation}
\theta^* = \arg\max_{\theta} \sum_{(\mathcal{V}, S)} \log p(S | \mathcal{V}, \theta).
\end{equation}

The final command combines outputs from both branches:
\begin{equation}
S = \mathcal{A} \oplus \mathcal{I}.
\end{equation}
where $\mathcal{A}$ is the predicted action and $\mathcal{I}$ represents the identified objects. For single-object actions, such as touch, pick, push, pull, and drop, only the target object is specified. For transfer actions, such as move, put, and pour, both the manipulated object and the target location are included. We use a grammar-free format, such as ``put apple into bowl,'' rather than a full natural-language sentence. This format emphasizes the essential components required for robotic execution.

\subsection{Action Representation Branch}~\label{sec:action_understanding}

The Action Representation Branch (ARB) aims to classify the manipulation primitive from the downsampled video $\tilde{\mathbf{F}}$ using a ResNet backbone augmented with Temporal Shift Modules (TSM)~\cite{lin2019tsm}. TSM shifts feature channels along the temporal dimension:
\begin{equation}
\tilde{X}_t^c = \begin{cases}
X_{t-1}^c & \text{if } c < C/4, \\
X_{t+1}^c & \text{if } C/4 \leq c < C/2, \\
X_t^c & \text{otherwise},
\end{cases}
\end{equation}
where $X_t^c$ denotes the $c$-th channel of features at time $t$, and $C$ is the total number of channels. This shifting enables information exchange among neighboring frames without additional computation or parameters. We implement TSM using the residual shift strategy by inserting the module into the residual branch rather than the main pathway. In-place shifting on the main pathway overwrites part of each layer's activations before convolution, which can weaken the spatial features used by the 2D backbone. By confining the shift to the residual branch, the identity path remains intact. The module classifies videos into the following manipulation primitives: \textit{touch}, \textit{pick}, \textit{move}, \textit{put}, \textit{open}, \textit{close}, \textit{push}, \textit{pull}, \textit{pour}, and \textit{drop}.

Given the temporally aggregated video representation $\mathbf{z}\in\mathbf{R}^{C}$, we predict an action distribution over $K$ primitives using a linear classifier $\mathbf{s} = W\mathbf{z} + \mathbf{b}$ followed by softmax. The classifier is trained with the cross-entropy loss:
\begin{equation}
L_{act}
=
-\sum_{k=1}^{K}
y_k \log \left( p_k \right),
\end{equation}
where $\mathbf{y}$ is the one-hot ground-truth action label.

\subsection{Object-Aware Branch}~\label{sec:object_understanding}

\subsubsection{Keyframe Extraction}

Frames captured during rapid motion are often degraded by motion blur, while frames from static periods may contain partially or fully occluded objects. Therefore, not all frames contribute equally to object identification. We select keyframes that retain informative visual changes for subsequent object recognition:
\begin{equation}
    \hat{\mathbf{F}} = f_t(f_s(T_G(\tilde{\mathbf{F}}))),
\end{equation}
where $T_G$ applies grayscale transformation, $f_s$ computes frame differences ($\tilde{f}_i - \tilde{f}_{i-1}$), and $f_t$ applies a threshold filter to retain significant visual changes. This process captures moments of meaningful interaction while filtering out static periods.

Concretely, we define a motion saliency score for frame $i$ as the average absolute pixel change:
\begin{equation}
\Delta_i = \frac{1}{HW}\sum_{u=1}^{H}\sum_{v=1}^{W} \left| T_G(\tilde{f}_i)(u,v) - T_G(\tilde{f}_{i-1})(u,v) \right|.
\end{equation}

Keyframes are then selected by thresholding and top-$M$ ranking:
\begin{equation}
\hat{\mathbf{F}}
=
\operatorname{TopM}
\left(
\{\tilde{f}_i\}_{i=2}^{m},
\Delta_i \mathbf{1}_{\{\Delta_i>\tau\}}
\right).
\end{equation}

\subsubsection{Object Detection, Tracking and Selection}
	
Keyframes $\hat{\mathbf{F}}$ are processed by a Human-Object Interaction (HOI) detector~\cite{shan2020understanding} to identify objects that interact with human hands. The detected objects are tracked across frames to generate trajectories. This produces a set of candidate objects $\mathbf{O} = \{o_1, o_2, \ldots, o_k\}$ with associated bounding boxes and motion trajectories.
A key contribution of our work is the Object Selection algorithm, which identifies the most relevant objects for command generation. It operates in two stages:

\textbf{Stage 1: Functional Role Classification.} 
The HOI detector identifies candidate interacting objects, but it does not determine which object is functionally primary when multiple detections appear across frames. We categorize objects into Pickable ($\mathbf{P}_1$) and Placeable ($\mathbf{P}_2$) sets using motion trajectory analysis. For transfer actions, such as pick, move, and put, the manipulated object usually shows greater displacement and vertical variation than stationary containers or surfaces. We use this relative asymmetry through a displacement threshold $\tau_D$ and a vertical-variance threshold $\tau_V$, without assuming a specific trajectory shape. Displacement alone cannot distinguish a lifted object from an object slid across a surface. Therefore, the vertical-variance term provides a height cue that separates these cases. For other primitives, such as push, pull, pour, and drop, object displacement patterns are less distinctive, making role separation less reliable. Thus, the Object Selection algorithm is mainly designed for transfer-type interactions.

For each tracked object $o$, let $\mathbf{c}_t(o)\in \mathbf{R}^2$ be the bounding-box center at time $t\in\{1,\dots,T\}$. We compute the total displacement and vertical variation as follows:
\begin{equation}
\begin{aligned}
    D(o)&=\sum_{t=2}^{T}\lVert \mathbf{c}_t(o)-\mathbf{c}_{t-1}(o)\rVert_2, \\ V(o)&=\operatorname{Var}\big(\{c_t^y(o)\}_{t=1}^{T}\big).
\end{aligned}
\end{equation}

We then assign functional roles using simple thresholds:
\begin{equation}
 o \in \mathbf{P}_1 \;\Leftrightarrow\; D(o) > \tau_D, V(o) > \tau_V,  o \in \mathbf{P}_2 \;\textit{otw},
\end{equation}
where, $\tau_D$ and $\tau_V$ are implemented as relative criteria derived from vertical motion and bounding-box evolution across tracked trajectories.

\textbf{Stage 2: Instance Quality Selection.} Within each functional category, we select the best instance using two complementary metrics. The \emph{blur score} $B(o)=\operatorname{Var}(\mathbf{I}*\mathbf{L})$ is the Laplacian-response variance of the cropped object image $\mathbf{I}$, where higher values indicate sharper crops. The \emph{hand overlap} $\operatorname{IoU}(o,h)=|B_o \cap B_h|/|B_o \cup B_h|$ is the intersection-over-union between the object box $B_o$ and the detected hand box $B_h$ in the same frame. We then choose
\begin{equation}
o^* = \arg\max_{o \in \mathbf{P}_i} \big[\alpha\, B(o) - \beta\, \operatorname{IoU}(o, h)\big],
\end{equation}
which maximizes visual sharpness while minimizing hand occlusion.
In our experiments, we set $\alpha = \beta = 1.0$, treating sharpness and occlusion as equally weighted. The Object Selection algorithm is designed for scenarios involving at most two functionally distinct objects. Performance degrades when three or more objects interact simultaneously, or when prolonged hand occlusion prevents reliable trajectory estimation.

\subsubsection{VLM-based Object Recognition}

Using selected object crops reduces irrelevant context and improves crop quality for object recognition. The selected crops are processed by a VLM~\cite{liu2023visual} to predict object categories. For each crop, we use the fixed prompt: ``What object is shown in this image? Respond with only the object name.'' The VLM output is parsed as a category label and assigned to the corresponding object instance. These labels are then used as object identifiers in the final command generation stage. Because object recognition is decoupled from action classification and relies on pretrained visual-semantic representations, the framework can recognize object categories that are not included in the training set.




\section{Experiments}~\label{sec:results}

\subsection{Experimental Setup}

\textbf{Dataset.} We train and evaluate our method on a modified Something-Something V2 dataset~\cite{goyal2017something}. We select 10 action categories that correspond directly to common robotic manipulation primitives. The training set contains 118,562 videos, and the test set contains 12,480 videos. We also collected 135 self-recorded videos with diverse backgrounds, lighting conditions, and object arrangements to evaluate real-world generalization. For object recognition, we define two evaluation sets: (1) \textit{Standard objects}: 12 categories present in training (apple, pan, bottle, orange, kettle, egg, plate, box, spoon, spatula, knife, cup); and (2) \textit{Novel objects}: 9 categories held out for zero-shot testing (carrot, block, grape, chili, banana, lemon, pot, strawberry, pressure cooker).
\begin{table}[!ht]
\centering
\caption{Action classification accuracy (\%) on benchmark test set and self-recorded videos.}
\label{tab:action}
\scriptsize
\begin{tabularx}{0.44\textwidth} { 
   >{\raggedright\arraybackslash} p{0.16\textwidth}
  || >{\centering\arraybackslash}X 
   >{\centering\arraybackslash}X  }
\hline
Test Set & Top-1 & Top-5 \\
\hline
Benchmark &  86.79 & 94.55 \\

Self-recorded & 68.15 & 94.81 \\
\hline
\end{tabularx}
\end{table}

\textbf{Implementation Details.} The ARB uses a ResNet-50 backbone with TSM. It is trained for 50 epochs using SGD with momentum 0.9, an initial learning rate of 0.01 with cosine annealing, and a batch size of 64. During inference, we sample 8 frames per video and apply 3-crop augmentation. We use the HOI detector from~\cite{shan2020understanding} for object detection and LLaVA~\cite{liu2023visual} with default parameters as our VLM. For a fair comparison, all baseline VLMs are strictly prompted to output only our grammar-free ground-truth format.


\textbf{Metrics.} Action classification is evaluated using Top-1 and Top-5 accuracy. Command generation quality is evaluated against ground-truth annotations using BLEU scores (B-1 through B-4), ROUGE-L (R), METEOR (M), and CIDEr (C). These metrics jointly measure lexical accuracy, sequence consistency, semantic similarity, and overall command relevance.

\subsection{Action Classification Results}

\begin{table}[!h]
\centering
\caption{Video-to-command generation performance.}
\label{tab:bleu_all}
\scriptsize
\begin{tabularx}{0.5\textwidth} { 
   >{\raggedright\arraybackslash} p{0.28\linewidth}
   >{\centering\arraybackslash}X 
   >{\centering\arraybackslash}X 
   >{\centering\arraybackslash}X 
   >{\centering\arraybackslash}X 
   >{\centering\arraybackslash}X 
   >{\centering\arraybackslash}X
   >{\centering\arraybackslash}X}
\hline
\multicolumn{8}{c}{a) Standard Object Sets} \\ \hline
\textbf{Method} & \textbf{B-1} & \textbf{B-2} & \textbf{B-3} & \textbf{B-4} & \textbf{R} & \textbf{M} & \textbf{C} \\
\hline

Video2Command\cite{nguyen2018translating}
& 0.312 & 0.196 & 0.173 & 0.150 & 0.330 & 0.234 & 1.103 \\

V2CNet\cite{nguyen2019v2cnet}
& 0.357 & 0.231 & 0.201 & 0.153 & 0.369 & 0.264 & 0.895 \\

Watch-and-Act\cite{yang2023watch} 
& 0.394 & 0.271 & 0.248 & 0.187 & 0.405 & 0.311 & 1.434 \\

ChatGPT-4o-mini\cite{achiam2023gpt} 
& 0.522 & 0.410 & 0.323 & \underline{0.269} & 0.574 & 0.478 & 1.636 \\

Qwen2.5-VL-3B\cite{wu2025qwen} 
& 0.531 & 0.322 & 0.254 & 0.207 & 0.547 & 0.428 & 1.307 \\

NVILA-8B\cite{liu2025nvila}
& 0.557 & 0.389 & 0.329 & 0.286 & 0.561 & 0.475 & \underline{1.773} \\

LLaVA-NeXT7B\cite{li2024llava}
& 0.280 & 0.171 & 0.105 & 0.075 & 0.347 & 0.234 & 0.493 \\

Intern2.5VL-4B\cite{chen2024expanding}  
& 0.525 & 0.319 & 0.244 & 0.172 & 0.511 & 0.429 & 1.078 \\

Intern3.5VL-4B\cite{wang2025internvl3}  
& \textbf{0.626} & \underline{0.427} & \underline{0.339} & 0.251 & \textbf{0.614} & \underline{0.521} & 1.422 \\

DeepseekVL7B\cite{lu2025deepseek} 
& 0.368 & 0.275 & 0.227 & 0.182 & 0.384 & 0.334 & 0.801 \\

\textbf{Ours (Full)}
& \underline{0.607} & \textbf{0.506} & \textbf{0.397} & \textbf{0.337}
& \underline{0.598} & \textbf{0.544} & \textbf{1.812} \\
\hline

\multicolumn{8}{c}{b) Novel Object Sets (Zero-shot)} \\ \hline

Video2Command\cite{nguyen2018translating}
& 0.305 & 0.150 & 0.090 & 0.082 & 0.328 & 0.203 & 0.467 \\

V2CNet\cite{nguyen2019v2cnet}
& 0.259 & 0.119 & 0.092 & 0.087 & 0.290 & 0.165 & 0.423 \\

Watch-and-Act\cite{yang2023watch} 
& 0.310 & 0.157 & 0.134 & 0.107 & 0.338 & 0.209 & 0.637 \\

ChatGPT-4o-mini\cite{achiam2023gpt} 
& \underline{0.444} & 0.338 & 0.240 & \underline{0.181} & \underline{0.485} & \underline{0.398} & 0.912 \\

Qwen2.5-VL-3B\cite{wu2025qwen} 
& 0.361 & 0.218 & 0.155 & 0.109 & 0.383 & 0.308 & 0.633 \\

NVILA-8B\cite{liu2025nvila}
& 0.387 & 0.245 & 0.161 & 0.110 & 0.400 & 0.365 & 0.669 \\

LLaVA-NeXT7B\cite{li2024llava}
& 0.159 & 0.084 & 0.060 & 0.047 & 0.202 & 0.118 & 0.190 \\

Intern2.5VL-4B\cite{chen2024expanding}  
& 0.347 & 0.199 & 0.135 & 0.092 & 0.374 & 0.298 & 0.551 \\

Intern3.5VL-4B\cite{wang2025internvl3}  
& 0.470 & \underline{0.341} & \underline{0.246} & \underline{0.181} & 0.474 & 0.414 & \underline{0.959} \\

DeepseekVL7B\cite{lu2025deepseek} 
& 0.329 & 0.235 & 0.166 & 0.127 & 0.343 & 0.306 & 0.558 \\

\textbf{Ours (Full)}
& \textbf{0.577} & \textbf{0.471} & \textbf{0.356} & \textbf{0.261}
& \textbf{0.558} & \textbf{0.539} & \textbf{1.731} \\
\hline

\end{tabularx}
\end{table}

Table~\ref{tab:action} presents the action classification results. On the benchmark test set, our TSM-based model achieves 86.79\% Top-1 accuracy and 94.55\% Top-5 accuracy. These results indicate reliable discrimination among fine-grained manipulation primitives, where temporal dynamics are crucial for distinguishing similar actions, such as ``pick'' and ``touch.''

On self-recorded videos, Top-1 accuracy decreases to 68.15\%. This decrease reflects appearance and viewpoint differences between curated benchmark videos and unconstrained real-world footage. However, Top-5 accuracy remains high at 94.81\%, indicating that the correct action class is usually among the top predictions even when the Top-1 prediction is incorrect.
\begin{figure}[!ht]
    \centering
    \includegraphics[width=\linewidth]{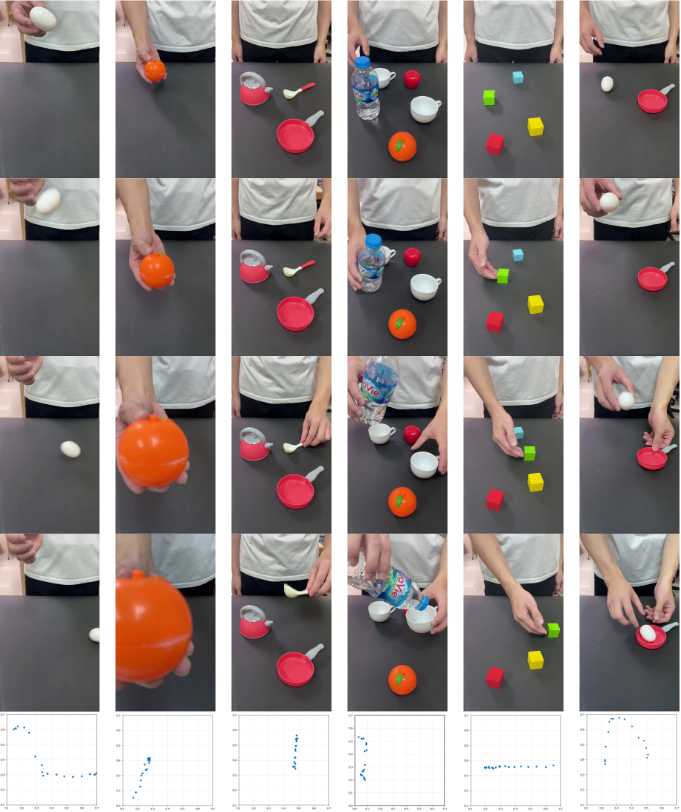}
    \caption{Qualitative examples of six manipulation actions with selected frames and object trajectories: (a) drop egg, (b) move orange, (c) pick spoon, (d) pour bottle into cup, (e) push block, (f) put egg into pan.}
    \label{fig:qualitative}
\end{figure}

\begin{figure*}[!ht]
    \centering
    \includegraphics[width=1\textwidth]{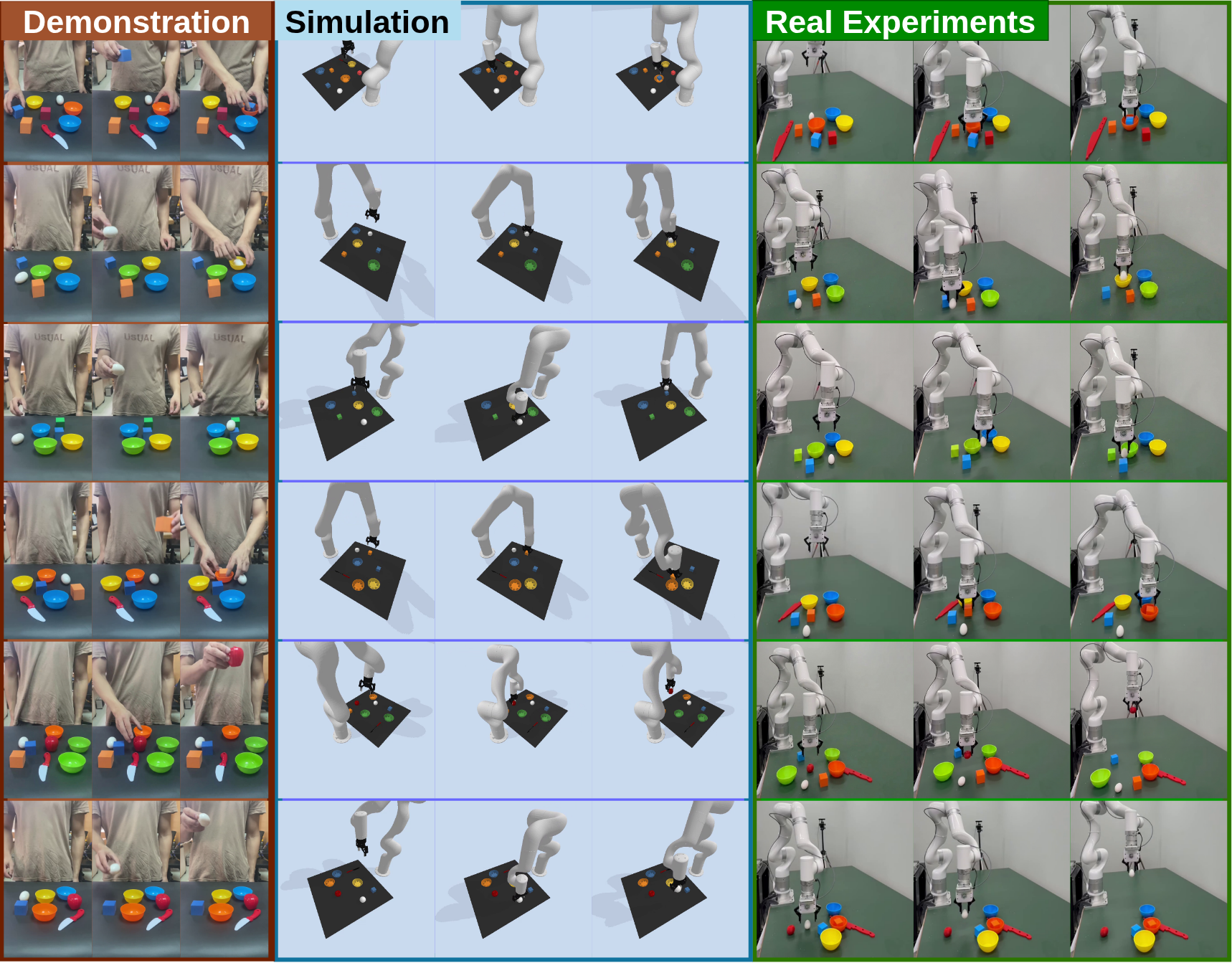}
    \caption{Qualitative examples of downstream robotic execution on six tasks.}
    \label{fig:real_demo}
\end{figure*}
\subsection{Video-to-Command Generation Results}

Table~\ref{tab:bleu_all} compares video-to-command generation performance. On standard objects (Table~\ref{tab:bleu_all}a), our method achieves a BLEU-4 score of 0.337. This outperforms the best task-specific baseline, Watch-and-Act (0.187), by 80.2\%.

The advantage becomes more pronounced on novel objects (Table~\ref{tab:bleu_all}b). Our method achieves a BLEU-4 score of 0.261, which is a 143.9\% improvement over Watch-and-Act (0.107). It also substantially outperforms all task-specific baselines on novel objects, despite using a compact grammar-free format and no task-specific fine-tuning. This result shows that VLM integration effectively uses pretrained semantic knowledge for zero-shot generalization, which is important for real-world deployment where robots encounter diverse objects.

Some large VLMs, such as LLaVA-NeXT7B, perform poorly on novel objects (B-4: 0.047). This suggests that end-to-end VLM inference on full video frames is unreliable for structured command generation when explicit object grounding is absent. We note that the baseline VLMs in Table~\ref{tab:bleu_all} operate on full video frames, which reflects their intended use as end-to-end video understanding systems. Therefore, we compare complete video-to-command pipelines rather than isolated object recognizers. ROUGE-L, METEOR, and CIDEr also support our approach. Although our method has a slightly lower ROUGE-L score on standard objects than InternVL3.5-4B (0.598 vs. 0.614), it achieves the best ROUGE-L score on novel objects (0.558 vs. 0.474), as well as the highest METEOR (0.544/0.539) and CIDEr (1.812/1.731) scores. Compared with Watch-and-Act, the gains reach 157.9\% in METEOR and 171.7\% in CIDEr on novel objects, indicating stronger command-level semantic alignment for novel categories.

\subsection{Qualitative Analysis}

Fig.~\ref{fig:qualitative} presents representative outputs across six action types. Compared with the baselines, our framework generates more specific commands by correctly identifying both actions and task-relevant objects. VLM integration also enables recognition of attributes such as color and material. Fig.~\ref{fig:real_demo} qualitatively shows that the generated action--object commands are compatible with downstream robotic execution interfaces.

\subsection{Ablation Studies}

Table~\ref{tab:ablation} presents the ablation results and validates the contribution of each component:

\begin{table}[!h]
\centering
\caption{Ablation study on framework components.}
\label{tab:ablation}
\scriptsize
\begin{tabularx}{0.46\textwidth} { 
   >{\raggedright\arraybackslash} p{0.2\textwidth}
  || >{\centering\arraybackslash}X 
   >{\centering\arraybackslash}X  }
\hline
Configuration & Standard & Novel \\
\hline
Full model & \textbf{0.337} & \textbf{0.261} \\ \hline
w/o keyframe extraction  & 0.298 & 0.218  \\
w/o object selection     & 0.237 & 0.193  \\
w/o VLM  & 0.312 & 0.142  \\
\hline
\end{tabularx}
\end{table}

\textbf{Keyframe Extraction:} Removing keyframe extraction and using uniform sampling reduces B-4 by 11.6\% on standard objects. Selecting interaction moments provides clearer object views than using arbitrary frames.

\textbf{Object Selection:} Removing the Object Selection algorithm and using all detected objects causes the largest degradation, with a 29.7\% decrease on standard objects. This confirms that trajectory analysis and quality filtering are essential for identifying task-relevant objects and generating precise commands.

\textbf{VLM vs. CNN Classifier:} Replacing the VLM with a trained CNN classifier reduces B-4 by 7.4\% on standard objects and 45.6\% on novel objects. This large gap on novel objects highlights the importance of the VLM's pretrained knowledge for zero-shot generalization.

\section{Conclusion}~\label{sec:conclusion}

In this paper, we presented a framework for video-to-command generation that separates temporal action recognition from object identification while preserving a unified command output. The Object Selection algorithm identifies task-relevant objects through trajectory-based role classification and quality filtering, while VLM integration provides zero-shot generalization to novel categories. Experimental results show consistent improvements over task-specific baselines across BLEU, METEOR, and CIDEr. The method also remains competitive with larger general-purpose VLMs without task-specific fine-tuning. The framework has several limitations. The Object Selection algorithm relies on clear spatial separation between objects and is restricted to two-object interactions. Its accuracy degrades under heavy overlap or sustained occlusion. Future work will address multi-object interactions using a learned trajectory classifier and will investigate real-time deployment.

\section*{Acknowledgment}
This work was supported by JST SPRING, Japan Grant Number JPMJSP2102.

\bibliographystyle{IEEEtran}
\bibliography{ref}

\end{document}